\documentclass[11pt]{article}

\usepackage[final]{acl}

\usepackage{times}
\usepackage{latexsym}
\usepackage{amsmath}
\usepackage{mathtools}
\usepackage{bm}
\usepackage{dsfont}
\usepackage{array}
\usepackage{tabularx}
\usepackage{booktabs}
\usepackage{multirow}
\usepackage[export]{adjustbox}
\usepackage{todonotes}
\usepackage{stfloats}

\usepackage[T1]{fontenc}

\usepackage[utf8]{inputenc}

\usepackage{microtype}

\usepackage{inconsolata}

\usepackage{graphicx}

\usepackage{xcolor}
\usepackage{pgf}
\usepackage{colortbl}

\definecolor{ACEgreen}{RGB}{220,245,220}
\definecolor{ACEblue}{RGB}{215,235,245}
\definecolor{ACEorange}{RGB}{245,235,210}
\definecolor{ACEred}{RGB}{245,215,215}

\newcommand{\ACEcell}[1]{%
  \pgfmathparse{#1}%
  \ifdim\pgfmathresult pt<0.05pt
    \cellcolor{ACEgreen}#1%
  \else\ifdim\pgfmathresult pt<0.10pt
    \cellcolor{ACEblue}#1%
  \else\ifdim\pgfmathresult pt<0.20pt
    \cellcolor{ACEorange}#1%
  \else
    \cellcolor{ACEred}#1%
  \fi\fi\fi
}

\newcommand{\BrierCell}[1]{%
  \pgfmathparse{#1}%
  \ifdim\pgfmathresult pt<0.10pt
    \cellcolor{ACEgreen}#1%
  \else\ifdim\pgfmathresult pt<0.20pt
    \cellcolor{ACEblue}#1%
  \else\ifdim\pgfmathresult pt<0.25pt
    \cellcolor{ACEorange}#1%
  \else
    \cellcolor{ACEred}#1%
  \fi\fi\fi
}

\title{Self-Aware Knowledge Probing: Evaluating Language Models' Relational Knowledge through Confidence Calibration
}

\author{Christopher Kissling\textsuperscript{1}, Elena Merdjanovska\textsuperscript{1,2} and Alan Akbik\textsuperscript{1,2} \\ 
 \textsuperscript{1}Humboldt-Universität zu Berlin \\ \textsuperscript{2}Science of Intelligence \\
 \texttt{c\_kissling@icloud.com}, \texttt{\{elena.merdjanovska, alan.akbik\}@hu-berlin.de}}
 
\begin{document}
\maketitle
\begin{abstract}
Knowledge probing quantifies how much relational knowledge a language model (LM) has acquired during pre-training. Existing knowledge probes evaluate model capabilities through metrics like prediction accuracy and precision. Such evaluations fail to account for the model's reliability, reflected in the calibration of its confidence scores. In this paper, we establish an empirical framework to evaluate the calibration of relational knowledge in a closed answer-set setting. We cover three modalities of model confidence: (1) intrinsic confidence, (2) paraphrase consistency and (3) linguistic confidence. Our extensive analysis of 21 causal and 8 masked language models reveals that most models, but especially those pre-trained with the masking objective, are overconfident. The best-calibrated scores come from  confidence estimates that account for inconsistencies due to statement rephrasing. Moreover, calibration improves when increasing the number of answer options. 

\begingroup
\renewcommand\thefootnote{}
\footnotetext{%
\includegraphics[height=2ex]{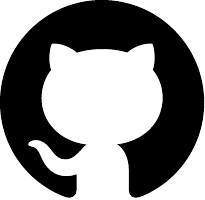}\hspace{0.6em}%
\url{https://github.com/diamir707/knowledge-calibration}%
}
\endgroup

\end{abstract}

\section{Introduction}
Knowledge probes serve as the primary diagnostic tool for quantifying the relational knowledge internalized by language models during pre-training. By querying models with factual triplets—typically formatted as cloze-style statements such as "Balach Marri died in [MASK]"—researchers can assess a model's grasp of diverse semantic relations \citep{petroni2019languagemodelsknowledgebases, youssef-etal-2023-give}. Such evaluations are critical for understanding how internalized knowledge correlates with performance on complex downstream reasoning tasks.

However, existing knowledge probing frameworks \citep{youssef-etal-2023-give, kalo2022kamel, wiland2024bearunifiedframeworkevaluating} focus almost exclusively on attainment: the accuracy and precision with which a model retrieves a fact. While accurate recall is imperative, it provides an incomplete picture of model reliability. A truly reliable model must be "self-aware", exhibiting confidence scores that are well-calibrated with its probability of being correct \citep{guo2017calibrationmodernneuralnetworks}. In a perfectly calibrated system, a confidence score of 0.8 implies an 80\% likelihood of factual correctness. Without such alignment, even high-accuracy models are prone to silent failures and overconfident hallucinations, as they lack the mechanisms to signal the limits of their own knowledge.

\begin{figure}[t!]
    \centering
    \includegraphics[width=1\linewidth]{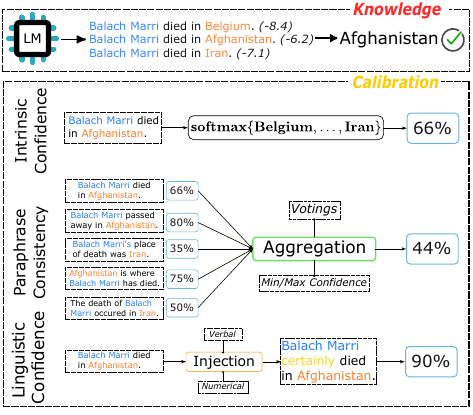}
    \caption{
    Using knowledge facts (like the place of death of a particular person), we measure an LM's awareness of its relational knowledge in three modalities: (1) intrinsic confidence, (2) paraphrase consistency of predictions across multiple rephrasings of the same statement, and (3) linguistic confidence using epistemic markers.
    }
    \label{fig:introduction}
\end{figure}

\noindent 
\textbf{Knowledge calibration probing.} 
In this paper, we empirically evaluate calibration techniques for relational knowledge. Unlike prior work that treats knowledge as a binary state of presence or absence, we evaluate the epistemic self-awareness of language models through calibration. We examine three distinct modalities of model confidence: (1)~\textit{intrinsic confidence}, derived from raw log-probabilities; (2)~\textit{paraphrase consistency}, measured through agreement across semantically equivalent statements; and (3)~\textit{linguistic confidence}, evaluated through the model's response to explicit linguistic markers of uncertainty (see \autoref{fig:introduction}).

This multi-faceted approach allows for a granular analysis of how model architecture and scale influence the reliability of factual retrieval. Specifically, we leverage a closed-set probing setup~\citep{wiland2024bearunifiedframeworkevaluating} to facilitate a direct, head-to-head comparison between causal language models (CLMs) and masked language models (MLMs). While these architectures differ fundamentally in their training objectives, 
our framework provides a unified environment to evaluate their respective epistemic self-awareness. To ensure our findings are robust across diverse knowledge domains and model families, we evaluate 23 distinct models ranging from 109M to 12B parameters. Among CLMs, we include base-variants and its corresponding instruction-tuned versions.

\noindent
\textbf{Contributions.} Our primary contributions include: 

\begin{itemize} 


\item We provide the first large-scale comparison of calibration between CLMs and MLMs, analyzing 23 models to isolate the impact of pre-training objectives on epistemic awareness.

\item We evaluate the efficacy of six confidence estimation strategies—ranging from raw log-probabilities to multi-prompt aggregations.

\item We investigate the impact of explicit linguistic confidence expressions on model reliability.
\end{itemize}

Our analysis reveals a critical "reliability gap" in current architectures; specifically, that MLMs—despite competitive factual accuracy—remain fundamentally less aware of their own knowledge than CLMs. Furthermore, we found a clear dependency between the number of distractors and calibration, showing that LMs need to be exposed to a sufficiently large number of answer options to achieve good calibration. 
Finally, our results suggest a clear scale-dependency for linguistic confidence: while smaller models are "semantically blind" to expressions of doubt, larger models begin to align their internal confidence with linguistic cues.
To facilitate further research into model reliability, we provide our proposed framework as an extensible benchmark for evaluating the epistemic awareness of future language models.

\section{Related Work}

The evaluation of model reliability in factual retrieval lies at the intersection of two research areas: uncertainty quantification for language models and relational knowledge probing.

\noindent 
\subsection{Confidence Estimation for LMs}
Confidence estimation methods can be broadly categorized into three modalities, mirroring the evaluation framework we propose in this paper.

\noindent 
\textbf{Intrinsic confidence.} The most direct approach utilizes the model’s internal probability distribution. Early work on calibration in neural networks \citep{guo2017calibrationmodernneuralnetworks} focused on softmax-normalized scores, a technique recently extended to LMs for tasks like hallucination detection \citep{huang2023lookbeforeleap} and selective prediction \citep{duan2024shiftingattentionrelevancepredictive}. Our work specifically examines how these intrinsic signals behave when the answer space is constrained to a fixed set of factual candidates.

\noindent 
\textbf{Paraphrase consistency.} This modality relies on the intuition that a model ``knows'' a fact if it provides the same answer across various semantically equivalent formulations. Techniques such as \textit{Self-Consistency} \citep{DBLP:conf/iclr/0002WSLCNCZ23} and \textit{Semantic Entropy} \citep{kuhn2023semantic} leverage multiple model outputs to derive uncertainty. While consistency is often used as a proxy for reliability in knowledge bases \citep{zheng2024reliablellmsknowledgebases}, our framework evaluates whether these consistency signals are well-calibrated with factual truth.

\noindent 
\textbf{Linguistic confidence.} A growing body of research explores ``self-reported'' or ``verbalized'' confidences, where the model expresses its certainty through natural language \citep{tian2023justaskcalibrationstrategies}. Previous studies have found that LMs are often reluctant to express uncertainty \citep{zhou2024relyingunreliableimpactlanguage} or fail to align their verbalized scores with their internal log-probabilities \citep{xiong2024llmsexpressuncertaintyempirical}. We extend this line of research to relational knowledge statements by injecting explicit epistemic markers (e.g., \textit{certainly}, \textit{possibly}) into the probing templates.

\subsection{Relational Knowledge Probing}
The field of knowledge probing was pioneered by the LAMA benchmark \citep{petroni2019languagemodelsknowledgebases}, which used cloze-style prompts to quantify the factual knowledge of MLMs. Subsequent benchmarks like KAMEL \citep{kalo2022kamel} and MyriadLAMA \citep{zhao-etal-2024-matters} expanded the diversity of relations and entities.

Benchmarks with cloze-style prompts provide a highly controlled environment to probe for relational knowledge. Unlike knowledge benchmarks in a question-answering format \citep{lin2022truthfulqameasuringmodelsmimic, hendrycks2021measuringmassivemultitasklanguage}, which introduce confounding variables like fluency or style \citep{rogers2023datasetexplosiontaxonomy}, cloze-style probes minimize linguistic uncertainties. This allows us to specifically isolate an LM's core factual capabilities. 


\subsection{The BEAR Framework}
\label{sec:bear-probe}
The BEAR probe \citep{wiland2024bearunifiedframeworkevaluating} establishes a unified knowledge probing mechanism for both CLMs and MLMs. 
Given a relation instance—such as (\textsc{Balach Marri}, \texttt{died-in}, \textsc{Afghanistan})—BEAR provides a set of $K$ candidate objects, comprising the true answer and plausible distractors (e.g. \textsc{Iran}). Using a relation-specific template, a statement is generated for each candidate—for instance, \emph{Balach Marri died in Afghanistan} and \emph{Balach Marri died in Iran}. The probe ultimately tests whether an LM can distinguish the true fact from the distractors by calculating the full log-likelihood of each statement.
\newline
\textbf{Scorings.} For CLMs, a sentence-level log-likelihood is simply the sum of the token log-likelihoods in the sequence:
\begin{equation*}
        \log P(x) = \sum_{t=1}^{|x|} \log P(x_t \mid x_{<t})
\end{equation*}
For MLMs, BEAR utilizes pseudo log-likelihoods (PLLs) \citep{Salazar_2020}. We adopt the refinement by \citet{kauf2023betterwaymaskedlanguage}, which masks tokens one-by-one and accounts for multi-token words by masking all future within-word tokens:
\begin{equation*}
        \text{PLL}(x) = \sum_{t=1}^{|x|} \log P(x_t \mid x_{\setminus t})
\end{equation*}

\noindent 
\textbf{Answer selection.} Once a log-likelihood score $\ell_i$ is assigned to every candidate statement $o_i$ in the answer set, the model's prediction $\hat{y}$ is the object that maximizes this score: $\hat{y} \coloneqq \text{arg}\max_{i \in \{1,\dots,K\}} \ell_i$. 

\noindent \textbf{Dataset Characteristics.} The dataset contains facts found in Wikidata, covering the following domains: \emph{Biographical}, \emph{Arts}, \emph{Geographic}, \emph{Movies}, \emph{Economic}, \emph{Historical}, \emph{Sports}, \emph{Scientific} and \emph{Political}. It should be noted that BEAR focuses on information prevalent enough to likely appear in pre-training corpora beyond Wikipedia. The resulting potential for test-set contamination is not necessarily a limitation; rather, for the purposes of knowledge probing, such exposure is expected. Finally, BEAR avoids key limitations of earlier probes, such as restrictions to single-subtoken answers, imbalanced answer distributions, and reliance on rare Wikidata entries. 

\section{Confidence Estimation}
In this section, we present the confidence estimates we calculated in order to assess calibration. For this, we first normalize the answer log-likelihoods using the softmax function ($\sigma$), yielding probabilities which can be interpreted as the model’s confidence in the correctness of a given answer (\autoref{fig:introduction}).

\subsection{Intrinsic Confidence}
Estimates in this category use a single template to calculate each instance’s log-likelihood.
\newline
\textbf{Base-Confidence.} We define a baseline estimate as the maximum value after softmax normalization:
\begin{equation}
	C_{\text{Base}}(\hat{y}) \coloneqq \max_{i \in \{1,\hdots,K\}} \hspace{0.3em} \sigma(\bm{\ell})_i
	\label{def:base_confidence}
\end{equation}
where $\bm{\ell}$ represents the vector of log-likelihoods of the answer options. We also used this baseline estimate to find the token log-likelihood reduction strategy which maximizes accuracy and minimizes calibration error (\autoref{sec:appendix-reduction-strategy}).
\newline
\textbf{Margin-Confidence.} Another single-template estimate is the confidence margin \cite{lin-etal-2022-towards}, also commonly used in the context of image classification \citep{pleiss2020identifyingmislabeleddatausing, joshi2009multiclassactivelearning}. With \emph{Margin-Confidence}, we compute the difference between the largest and second-largest value returned by the softmax, reflecting whether the model's answer is clearly preferred over the second-best answer:
\begin{equation}
	C_{\text{Margin}}(\hat{y}) \coloneqq \sigma(\boldsymbol{\ell})_{[1]} - \sigma(\boldsymbol{\ell})_{[2]}
	\label{def:margin_confidence}
\end{equation}
Here, the subscripts $[1]$ and $[2]$ refer to the largest and second-largest values returned by the softmax.
\subsection{Paraphrase Consistency}
\label{sec:paraphrase_consistency}
Paraphrase consistency estimates rely on multiple rephrasings of each relation-specific template.\footnote{Different from consistency often mentioned in knowledge editing literature \citep{dong-etal-2025-chainedit}, where it refers to logical consistency.} 
Repeating the probe across the templates yields multiple answers $\tilde{y}_1, \dots, \tilde{y}_5$, requiring an aggregation step to obtain a single prediction $\hat{y}$ (\autoref{fig:introduction}).
\newline
\textbf{Answer Aggregation.} The answers can be aggregated through voting schemes or confidence-based aggregation. Voting schemes require a minimum level of agreement (for example, at least 2 answers need to be the same). If agreement is insufficient, the prediction is rejected, yielding no answer and zero calibration error by design. Confidence-based aggregation always produces a final answer, which in some cases can rely only on a single candidate.
In our experiments, we use confidence-based aggregation, selecting the predicted answer with minimum $C_{\text{Base}}$.

\noindent
\textbf{Average-Confidence.} We define \emph{Average-Confidence} as the per-template average of the selected answer:
\begin{equation}
	C_{\text{Average}}(\hat{y})\coloneqq \frac{\sum_{i=1}^5 \mathds{1}(\tilde{y}_i=\hat{y})\cdot C_{\text{Base}}(\tilde{y}_i)}{5}
    \label{def:average_confidence}
\end{equation}

\noindent
\textbf{Consistency-Confidence.} Instead of averaging the softmax scores, confidence can be reflected through the relative frequency of the answers:
\begin{equation}
	C_{\text{Consistency}}(\hat{y})\coloneqq \frac{\sum_{i=1}^5 \mathds{1}(\tilde{y}_i=\hat{y})}{5}
    \label{def:consistency_confidence}
\end{equation}
Hence, $C_{\text{Consistency}}$ bypasses the step of softmax normalization and is discretely distributed. It therefore more closely aligns with how humans naturally express confidence, typically in multiples of five \cite{xiong2024llmsexpressuncertaintyempirical}. 

\noindent
\textbf{Entropy-Confidence.} The certainty of the predicted answer can also be measured through the entropy over the predicted answer distribution. For each unique predicted answer $\tilde{y}_j \in \tilde{Y}$, we compute the unnormalized average $C_{\text{Base}}$:

\begin{equation}
    s(\tilde{y}_j)\coloneqq \frac{\sum_{i=1}^{5} \mathds{1}(\tilde{y}_i = \tilde{y}_j) \cdot C_{\text{Base}}(\tilde{y}_i)}{\sum_{i=1}^{5} \mathds{1}(\tilde{y}_i = \tilde{y}_j)}
\end{equation}

\noindent
Then, we normalize it to obtain a confidence distribution: 

\begin{equation}
    q(\tilde{y}_j)\coloneqq\frac{s(\tilde{y}_j)}
    {
        \sum_{\tilde{y}_k \in \tilde{Y}}s(\tilde{y}_k)
    }
\end{equation}

\noindent
Since the Shannon entropy is a measure of certainty, we subtract it from 1 to obtain our entropy confidence estimate:
\begin{equation}
	C_{\text{Entropy}}(\hat{y})\coloneqq 1 -  \Big(-\sum_{\tilde{y}_j \in \tilde{Y}} q(\tilde{y}_j) \log q(\tilde{y}_j)\Big)
    \label{def:entropy_confidence}
\end{equation}

We base this approach on the widely-used semantic entropy \citep{kuhn2023semantic}. In our approach, the semantic clusters correspond to the predicted unique answers. Hence, with five templates, there is a maximum of five clusters per instance. 
\subsection{Linguistic Confidence}
To assess linguistic confidence calibration, we evaluate the impact of \emph{verbalized} and \emph{numerical} confidence expressions 
on both accuracy and calibration. To achieve this, we modify the templates by injecting verbalized and numerical confidence expressions \cite{zhou2023navigatinggreyareaexpressions}. As a confidence score, we use the single-template $C_{\text{Base}}$ (\autoref{def:base_confidence}).
\newline
\textbf{Verbalized.} For the verbalized setup, we inject the epistemic marker \emph{possibly} to lower the confidence of the statement and the marker \emph{certainly} to strengthen it. For instance, \emph{Balach Marri certainly died in Afghanistan}.
\newline
\textbf{Numerical.} To evaluate the impact of numerical confidence expressions in relational knowledge statements, we modified the relation templates to include formulations such as \emph{“I’m $x$\% confident that Balach Marri died in Afghanistan.”}, with $x \in \{0, 25, \ldots, 100\}$.

\section{Experimental Setup}
In this section, we give an overview of the datasets, models and evaluation metrics we used. 

\subsection{Datasets}

We base our calibration probe around the BEAR framework \citep{wiland2024bearunifiedframeworkevaluating}. The dataset contains 7,731 instances across 60 relations. Each relation has three semantically equivalent template statements. As part of this work, we extended the number of templates to five per relation using ChatGPT (see \autoref{sec:appendix-extending-templates}). Out of all relations, 46 have cardinality N:1 (different subjects can share the same answer) and 14 have cardinality 1:1 (each object is the answer to exactly one subject). For the N:1 relations, the average number of answer options per instance is 6.5. Each 1:1 relation consists of 60 instances, with 60 answer options per instance.

Additionally, we run experiments with the Wiki subset of the FACTOR \cite{muhlgay-etal-2024-generating} dataset, which has one correct and three incorrect completions for each statement. 

\subsection{Language Models}
We evaluated 16 different models, 10 CLMs and 6 MLMs (listed in \autoref{sec:appendix-models}). The MLMs range from a size of 109 million parameters to 561 million parameters. For the CLMs, the smallest model is the \texttt{opt-125m} with 125 million parameters, and the largest is \texttt{gemma-12b(IT)} with 12 billion parameters. We include examples of both base and instruction-tuned CLMs.

\begin{table*}[htp!]
    \centering
    \small
    \setlength{\tabcolsep}{6pt}
\begin{adjustbox}{width=\textwidth}
\begin{tabular}{lccccccccc}
\toprule & \multicolumn{4}{c}{\textbf{ACE ($\downarrow$)}} & \multicolumn{5}{c}{\textbf{Brier Score ($\downarrow$)}}\\ \cmidrule(lr){2-5} \cmidrule(lr){6-10}

\textbf{Model} 
& $C_{\text{Base}}$ 
& $C_{\text{Margin}}$ 
& $C_{\text{Average}}$ 
& $C_{\text{Entropy}}$ 
& $C_{\text{Base}}$ 
& $C_{\text{Margin}}$ 
& $C_{\text{Average}}$ 
& $C_{\text{Consistency}}$
& $C_{\text{Entropy}}$  \\
\midrule
\texttt{gemma-2b} & \ACEcell{0.113} & \ACEcell{0.053} & \textbf{\ACEcell{0.032}} & \ACEcell{0.150} & \BrierCell{0.162} & \BrierCell{0.155} & \textbf{\BrierCell{0.098}} & \BrierCell{0.157} & \BrierCell{0.155} \\
\texttt{gemma-7b} & \ACEcell{0.077} & \ACEcell{0.046} & \textbf{\ACEcell{0.042}} & \ACEcell{0.172}& \BrierCell{0.133} & \BrierCell{0.134} & \textbf{\BrierCell{0.083}} & \BrierCell{0.124} & \BrierCell{0.144} \\
\midrule
\texttt{opt-125m}
& \ACEcell{0.299} & \ACEcell{0.153} & \ACEcell{0.116} & \textbf{\ACEcell{0.106}} & \BrierCell{0.241} & \BrierCell{0.186} & \textbf{\BrierCell{0.130}} & \BrierCell{0.273} & \BrierCell{0.159} \\

\texttt{opt-350m} 
& \ACEcell{0.265} & \ACEcell{0.139} & \textbf{\ACEcell{0.098}} & \ACEcell{0.109} & \BrierCell{0.224} & \BrierCell{0.177} & \textbf{\BrierCell{0.127}} & \BrierCell{0.263} & \BrierCell{0.166} \\

\texttt{opt-1.3b} 
& \ACEcell{0.207} & \ACEcell{0.100} & \textbf{\ACEcell{0.056}} & \ACEcell{0.118} & \BrierCell{0.214} & \BrierCell{0.187} & \textbf{\BrierCell{0.126}} & \BrierCell{0.232} & \BrierCell{0.177} \\

\texttt{opt-6.7b} 
& \ACEcell{0.154} & \ACEcell{0.069} & \textbf{\ACEcell{0.024}} & \ACEcell{0.138} & \BrierCell{0.187} & \BrierCell{0.168} & \textbf{\BrierCell{0.110}} & \BrierCell{0.191} & \BrierCell{0.163} \\

\midrule

\texttt{gemma-3-270m (IT)} & \ACEcell{0.395} & \ACEcell{0.235} & \textbf{\ACEcell{0.125}} & \ACEcell{0.128} & \BrierCell{0.303} & \BrierCell{0.218} & \textbf{\BrierCell{0.116}} & \BrierCell{0.220} & \BrierCell{0.165} \\
\texttt{gemma-3-1b (IT)} & \ACEcell{0.350} & \ACEcell{0.210} & \textbf{\ACEcell{0.092}} & \ACEcell{0.198} & \BrierCell{0.284} & \BrierCell{0.217} & \textbf{\BrierCell{0.108}} & \BrierCell{0.187} & \BrierCell{0.226} \\
\texttt{gemma-3-4b (IT)} & \ACEcell{0.267} & \ACEcell{0.162} & \textbf{\ACEcell{0.053}} & \ACEcell{0.338}  & \BrierCell{0.240} & \BrierCell{0.193} & \textbf{\BrierCell{0.099}} & \BrierCell{0.147} & \BrierCell{0.329}  \\
\texttt{gemma-3-12b (IT)} & \ACEcell{0.193} & \ACEcell{0.117} & \textbf{\ACEcell{0.053}} & \ACEcell{0.436} & \BrierCell{0.185} & \BrierCell{0.154} & \textbf{\BrierCell{0.084}} & \BrierCell{0.117} & \BrierCell{0.370}  \\
\midrule[0.3pt]
\texttt{Llama-3.2-1B (IT)} & \ACEcell{0.170} & \ACEcell{0.072} & \textbf{\ACEcell{0.036}} & \ACEcell{0.288}& \BrierCell{0.190} & \BrierCell{0.167} & \textbf{\BrierCell{0.110}} & \BrierCell{0.189} & \BrierCell{0.290} \\
\texttt{Llama-3.2-3B (IT)} & \ACEcell{0.129} & \ACEcell{0.057} & \textbf{\ACEcell{0.039}} & \ACEcell{0.367}  & \BrierCell{0.167} & \BrierCell{0.153} & \textbf{\BrierCell{0.098}} & \BrierCell{0.150} & \BrierCell{0.329} \\
\texttt{Llama-3.1-8B (IT)} & \ACEcell{0.086} & \textbf{\ACEcell{0.040}} & \ACEcell{0.048} & \ACEcell{0.433} & \BrierCell{0.130} & \BrierCell{0.126} & \textbf{\BrierCell{0.076}} & \BrierCell{0.110} & \BrierCell{0.316} \\
\midrule[0.3pt]
\texttt{Qwen3.5-0.8B (IT)} & \ACEcell{0.247} & \ACEcell{0.118} & \textbf{\ACEcell{0.083}} & \ACEcell{0.162} & \BrierCell{0.218} & \BrierCell{0.176} & \textbf{\BrierCell{0.113}} & \BrierCell{0.247} & \BrierCell{0.204} \\
\texttt{Qwen3.5-2B (IT)} & \ACEcell{0.203} & \ACEcell{0.087} & \textbf{\ACEcell{0.049}} & \ACEcell{0.218} & \BrierCell{0.201} & \BrierCell{0.172} & \textbf{\BrierCell{0.112}} & \BrierCell{0.222} & \BrierCell{0.250}  \\
\texttt{Qwen3.5-4B (IT)} & \ACEcell{0.152} & \ACEcell{0.058} & \textbf{\ACEcell{0.040}} & \ACEcell{0.297}  & \BrierCell{0.172} & \BrierCell{0.153} & \textbf{\BrierCell{0.098}} & \BrierCell{0.165} & \BrierCell{0.283}  \\
\texttt{Qwen3.5-9B (IT)} & \ACEcell{0.114} & \ACEcell{0.047} & \textbf{\ACEcell{0.034}} & \ACEcell{0.371}  & \BrierCell{0.155} & \BrierCell{0.146} & \textbf{\BrierCell{0.095}} & \BrierCell{0.153} & \BrierCell{0.320} \\

\midrule[0.3pt]

\texttt{bert-base-cased} 
& \ACEcell{0.454} & \ACEcell{0.295} & \ACEcell{0.218} & \textbf{\ACEcell{0.129}} & \BrierCell{0.367} & \BrierCell{0.276} & \textbf{\BrierCell{0.205}} & \BrierCell{0.338} & \BrierCell{0.247} \\

\texttt{bert-large-cased} 
& \ACEcell{0.452} & \ACEcell{0.302} & \ACEcell{0.198} & \textbf{\ACEcell{0.125}} & \BrierCell{0.373} & \BrierCell{0.285} & \textbf{\BrierCell{0.196}} & \BrierCell{0.310} & \BrierCell{0.223} \\

\midrule[0.3pt]

\texttt{roberta-base} 
& \ACEcell{0.482} & \ACEcell{0.321} & \ACEcell{0.165} & \textbf{\ACEcell{0.115}} & \BrierCell{0.387} & \BrierCell{0.291} & \textbf{\BrierCell{0.162}} & \BrierCell{0.265} & \BrierCell{0.174} \\

\texttt{roberta-large} 
& \ACEcell{0.441} & \ACEcell{0.291} & \ACEcell{0.144} & \textbf{\ACEcell{0.124}} & \BrierCell{0.367} & \BrierCell{0.278} & \textbf{\BrierCell{0.158}} & \BrierCell{0.254} & \BrierCell{0.183} \\

\midrule[0.3pt]

\texttt{xlm-roberta-base} 
& \ACEcell{0.585} & \ACEcell{0.427} & \ACEcell{0.274} & \textbf{\ACEcell{0.104}} & \BrierCell{0.481} & \BrierCell{0.368} & \textbf{\BrierCell{0.225}} & \BrierCell{0.362} & \BrierCell{0.246} \\

\texttt{xlm-roberta-large} 
& \ACEcell{0.559} & \ACEcell{0.412} & \ACEcell{0.249} & \textbf{\ACEcell{0.119}} & \BrierCell{0.461} & \BrierCell{0.359} & \textbf{\BrierCell{0.224}} & \BrierCell{0.354} & \BrierCell{0.259} \\

\midrule[0.3pt]

\texttt{ModernBERT-base} & \ACEcell{0.462} & \ACEcell{0.303} & \ACEcell{0.143} & \textbf{\ACEcell{0.128}} & \BrierCell{0.372} & \BrierCell{0.279} & \textbf{\BrierCell{0.147}} & \BrierCell{0.248} & \BrierCell{0.194} \\

\texttt{ModernBERT-large} & \ACEcell{0.400} & \ACEcell{0.254} & \textbf{\ACEcell{0.116}} & \ACEcell{0.153} & \BrierCell{0.332} & \BrierCell{0.255} & \textbf{\BrierCell{0.151}} & \BrierCell{0.242} & \BrierCell{0.220} \\
\bottomrule
\end{tabular}
\end{adjustbox}
\caption{\emph{ACE} and \emph{Brier Score} of our estimates.
\textbf{Bold values} indicate the best-performing estimate per model.
\textbf{\emph{ACE:}}
\protect\textcolor[rgb]{0.35,0.60,0.35}{\scalebox{1.5}{\textbullet}}~[0,0.05),
\protect\textcolor[rgb]{0.30,0.50,0.65}{\scalebox{1.5}{\textbullet}}~[0.05,0.10),
\protect\textcolor[rgb]{0.65,0.50,0.25}{\scalebox{1.5}{\textbullet}}~[0.10,0.20),
\protect\textcolor[rgb]{0.65,0.30,0.30}{\scalebox{1.5}{\textbullet}}~$\geq 0.20$;
\textbf{\emph{Brier Score:}}
\protect\textcolor[rgb]{0.35,0.60,0.35}{\scalebox{1.5}{\textbullet}}~[0–0.10),
\protect\textcolor[rgb]{0.30,0.50,0.65}{\scalebox{1.5}{\textbullet}}~[0.10–0.20),
\protect\textcolor[rgb]{0.65,0.50,0.25}{\scalebox{1.5}{\textbullet}}~[0.20–0.25),
\protect\textcolor[rgb]{0.65,0.30,0.30}{\scalebox{1.5}{\textbullet}}~$\geq 0.25$. A \emph{Brier Score} of 0.25 corresponds to a \emph{random baseline}, always predicting a confidence of 0.5. For \emph{ACE}, we consider values < 0.1 \emph{well-calibrated}, where the confidence deviates from accuracy by 10 percentage points at most. We do not report \emph{ACE} for $C_{\text{Consistency}}$ due to its discrete distribution.}
    \label{tab:ace_brier_conf_eval_updated}
\end{table*}

\subsection{Evaluation Metrics}
\textbf{Adaptive Calibration Error.} Due to the limitations of the commonly-used Expected Calibration Error (ECE) \cite{naeini2015BBQcalibratedprobabilities} with fixed bin widths, we use the \emph{Adaptive Calibration Error} (\emph{ACE}) \cite{nixon2020measuringcalibrationdeeplearning}. ACE accounts for the underlying confidence distributions by defining adaptive bin intervals so that each contains an equal number of predictions. Low \emph{ACEs} indicate high knowledge awareness. 
\newline
\textbf{Calibration Curves.} To assess over- and under-confidence, we rely on calibration curves for which we plot the confidence scores against accuracy, i.e. the fractions of correct predictions in each 
confidence bin. For both calculating \emph{ACE} and plotting the calibration curves, we use quantile binning with 20 bins.
\newline
\textbf{Brier Score.} We include the \emph{Brier Score} \cite{brier1950forecastverification} to reflect both calibration and discriminative power. Unlike ACE, which would be zero for a random guessing binary classifier that always predicts 50\% confidence, the Brier Score penalizes such uninformative estimates \cite{tian2023justaskcalibrationstrategies}. We use correctness labels $\mathds{1}(y_i = \hat{y}_i)$ as a target \cite{tian2023justaskcalibrationstrategies, lin2022teachingmodelsexpressuncertainty}. 
\newline

\section{Experiment 1: Calibration Evaluation}
This experiment evaluates the calibration of the confidence estimates for each model. 
\subsection{Comparison of Confidence Estimates}
\begin{figure}[tb!]
	\center
	\includegraphics[width=1\linewidth]{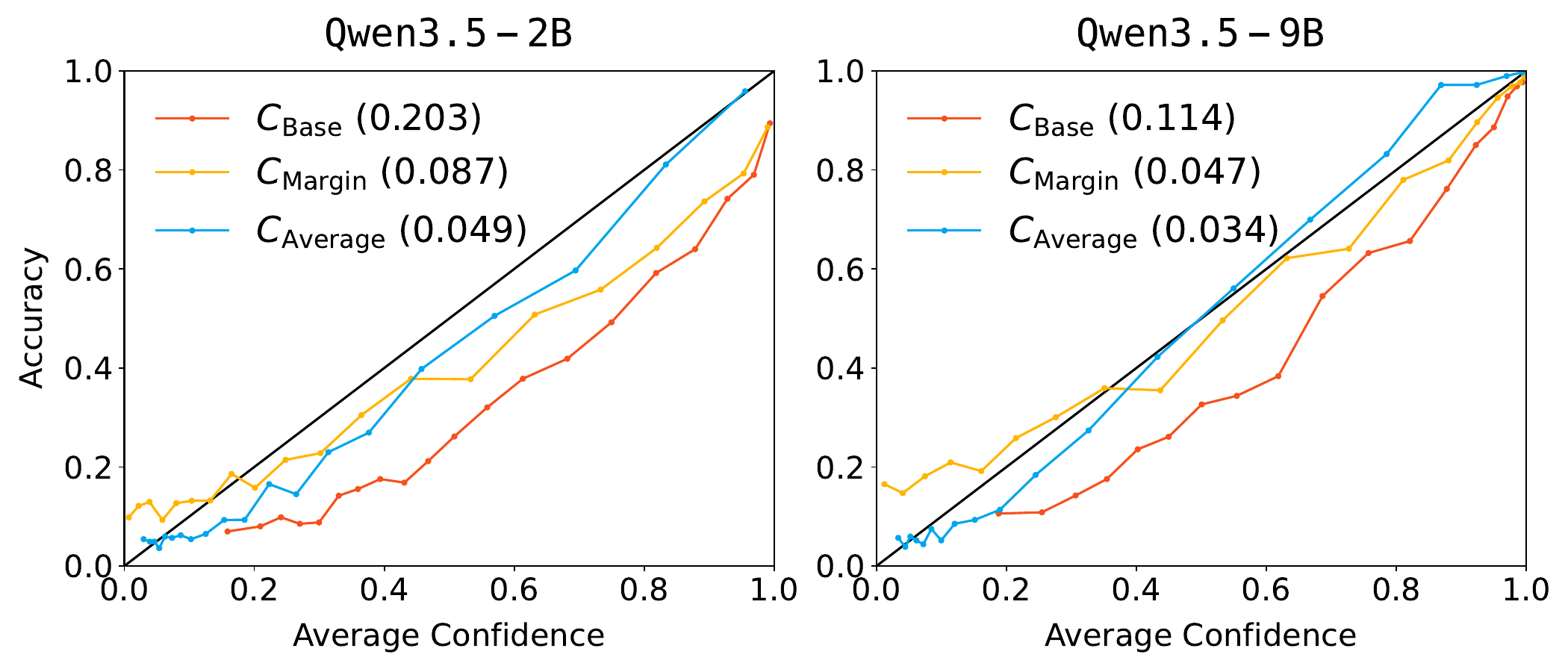}
	\caption{Calibration curves of our continuous confidence estimates. \emph{ACE} in parentheses.}
	\label{fig:calibration_curves_estimates}
\end{figure}

\autoref{tab:ace_brier_conf_eval_updated} presents the \emph{ACEs} and the \emph{Brier Scores} for all models and each confidence estimate. \autoref{fig:calibration_curves_estimates} shows the calibration curves for selected models.

\noindent $\boldsymbol{C_{\text{Margin}}}$ \textbf{scores are more reliable than} $\boldsymbol{C_{\text{Base}}.}$
From the intrinsic confidence estimates, derived using only a single template, $C_{\text{Margin}}$ emerged as a consistently well-calibrated method 
across all LMs. 
With $C_{\text{Margin}}$, \texttt{gemma-7b}, \texttt{Llama-3.1-8B(IT)} and \texttt{Qwen3.5-9B(IT)} achieve an \emph{ACE} of under 0.05, indicating good calibration and reliability. 
This efficacy is further illustrated by the calibration curves in \autoref{fig:calibration_curves_estimates}, where we see that $C_{\text{Base}}$ is markedly overconfident. On the other hand, the $C_{\text{Margin}}$ curves align more closely with the identity line, which signifies perfect calibration.

For all models included in \autoref{fig:calibration_curves_estimates}, $C_{\text{Base}}$ does not produce enough scores in the lower confidence ranges to derive a calibration curve 
that covers the full $[0,1]$ interval. 
$C_{\text{Margin}}$, however, does extend the curve into these lower confidence regions without limiting the model's ability to output enough high-confidences, allowing for evaluation across the entire range. We also observe that $C_{\text{Margin}}$ can introduce underconfidence in the lower range of the confidence scores, pushing the calibration curve above the identity line. From a practical perspective, an underconfident model is generally preferable than an overconfident one.

We confirm the benefits of $C_{\text{Margin}}$ through the Wiki-FACTOR dataset. In \autoref{tab:ace_brier_factor} in \autoref{sec:factor-benchmark}, we see that $C_{\text{Margin}}$ consistently outperforms $C_{\text{Base}}$ across all models and both evaluation metrics, highlighting it as a more reliable confidence estimate.

\noindent $\boldsymbol{C_{\text{Average}}^{}}$ \textbf{is the best-calibrated estimate.} The paraphrase consistency estimate 
$C_{\text{Average}}^{}$ clearly outperforms the other approaches across all models in terms of both \emph{Brier Score}  and \emph{ACE}. For example, for \texttt{opt-6.7b}, $C_{\text{Average}}^{}$ achieves an \emph{ACE} of 0.024, and for \texttt{Qwen3.5-9B(IT)}, an \emph{ACE} of 0.034, which is close to perfect calibration. This is reflected in the calibration curve shown in \autoref{fig:calibration_curves_estimates}, which deviates only slightly from the identity line. We further note that for larger models (for example, \texttt{gemma-7b} and \texttt{Qwen3.5-9B(IT)}), the difference between $C_{\text{Average}}$ and $C_{\text{Margin}}$ is often small, with $C_{\text{Margin}}$ providing better-aligned scores than $C_{\text{Average}}$ for \texttt{Llama-3.1-8B(IT)}. 

The other paraphrase consistency estimates: $C_{\text{Consistency}}$ and $C_{\text{Entropy}}$ provide very mis-calibrated scores with low discriminative power
, showing that simple answer agreement is not enough to capture model confidence. 

\noindent \textbf{The largest models are also most calibrated.} For most model families, we see that by increasing model size, \emph{ACE}s and \emph{Brier Scores} decrease. This applies to all confidence estimation methods. 
The model families with the overall lowest \emph{ACE} and \emph{Brier Scores} are \texttt{LlaMA} and \texttt{gemma}, which indicates that these models' confidences best reflect their factual knowledge. Newer \texttt{gemma-3} models exhibit lower accuracy (see \autoref{sec:appendix-models}) and worse calibration than the original \texttt{gemma} family. In general, all models larger than 1B parameters, achieve an $\emph{ACE} \leq 0.1$ with their best estimate, which we consider well-calibrated. Among the MLMs, we found that the \texttt{ModernBERT}'s $C_\text{Base}$ confidences are comparable to \texttt{BERT} and \texttt{RoBERTa}, however with $C_\text{Average}$ \texttt{ModernBERT} is better-calibrated. Overall, \texttt{xlm-roberta-large} deviates from the other MLMs with worse accuracy and calibration, demonstrating the negative impact of multi-lingual training data. 


\subsection{Impact of Answer Set Size 
}

\begin{figure}[t!]
    \centering
    \includegraphics[width=1\linewidth]{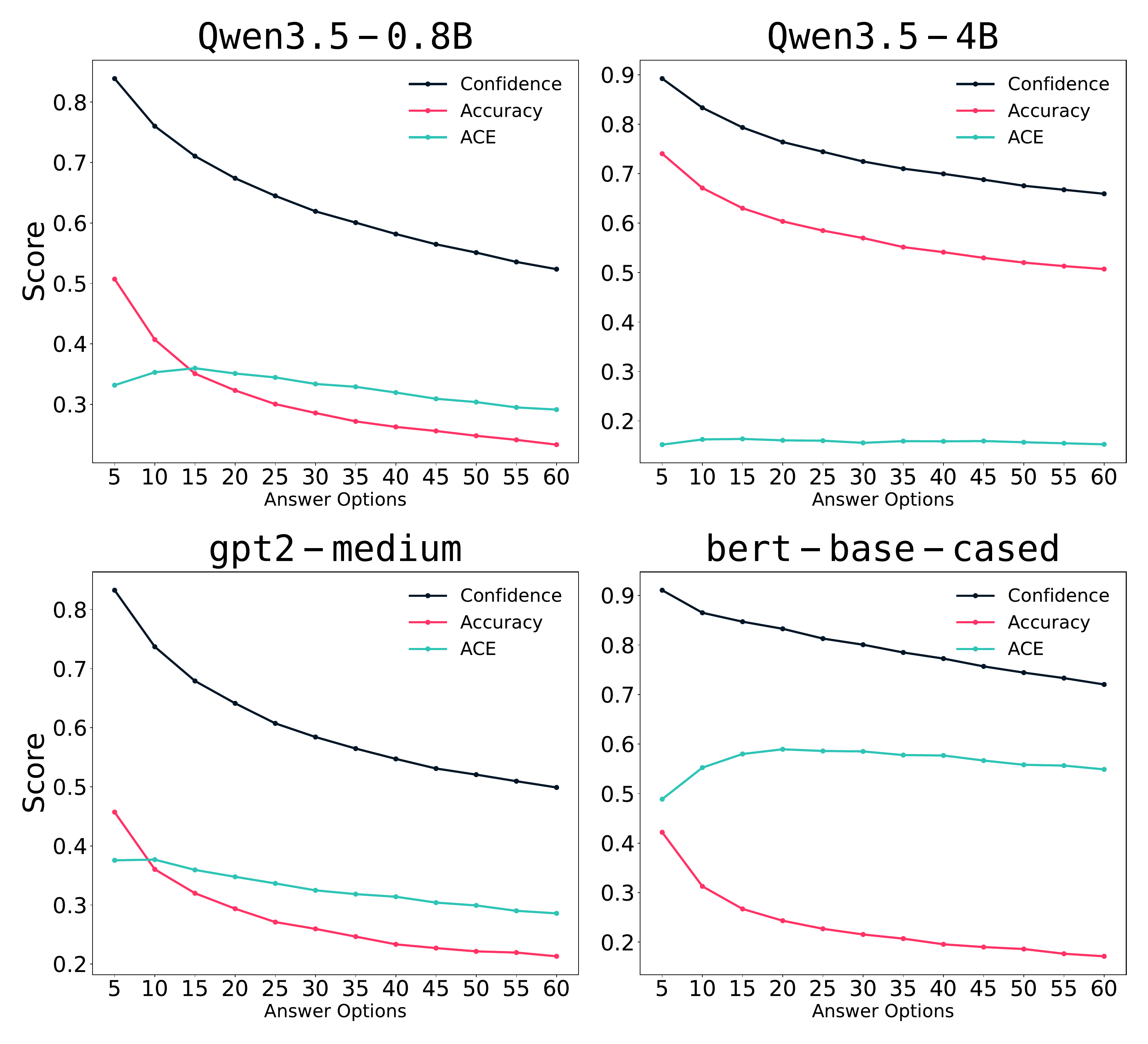}
    \caption{Average $C_{\text{Base}}$ confidence, accuracy and \emph{ACE} as a function of the number of answer options (including the correct answer). Each point represents the mean over three repeated samplings.}
    \label{fig:metrics_sample_size}
\end{figure}

This section investigates the influence of distractor count on intrinsic confidence. We analyze the average $C_{\text{Base}}$ confidence, accuracy and \emph{ACE} as a function of the number of answer options, sampled as subsets of the 60 options in the 1:1 relations in BEAR. \autoref{fig:metrics_sample_size} shows these relationships for selected models, and we observed similar patterns for the other models of each type. 

\noindent \textbf{More answer options lead to better calibration.}
We find that increasing the number of answer options (i.e. more incorrect answers as distractors)  significantly reduces calibration error in CLMs smaller than 1B (see \texttt{Qwen3.5-0.8B} and \texttt{gpt2-medium} in \autoref{fig:metrics_sample_size}). 
While both average confidence and accuracy decline as options increase, they do so at different rates: confidence drops faster than accuracy, reducing \emph{ACE}. 
In larger CLMs and all MLMs this effect is less pronounced, but still present. For \texttt{Qwen3.5-4B} and \texttt{bert-base-cased}, accuracy drops faster than confidence at low answer counts, thereby increasing \emph{ACE}. After 15--20 answer options, \emph{ACE} stabilizes, with no further calibration gains for larger CLMs and MLMs. Therefore, robust confidence estimation for both model types requires a fairly high number of distractors, albeit at a higher computational cost. 

\begin{figure}[tb!]
    \centering
    \includegraphics[width=0.7\linewidth]{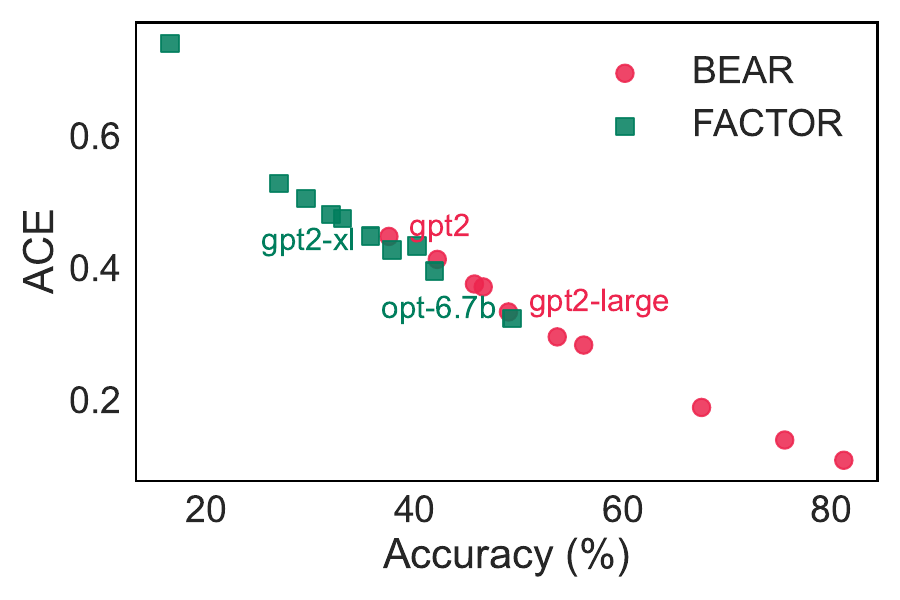}
    \caption{$\emph{Accuracy}$ against $\emph{ACE}$ of $C_{\text{Base}}$ for the 1:1 relations of BEAR and the Wiki-FACTOR dataset. For BEAR, we sampled the number of answer options to five per instance. On FACTOR, each question consists of four answer options.}
    \label{fig:ace_vs_acc_bear_factor}
\end{figure}

We confirm these findings through the Wiki-FACTOR benchmark. While FACTOR initially exhibits 
significantly higher calibration errors than BEAR (see Table \autoref{tab:ace_brier_factor} in \autoref{sec:factor-benchmark}), we primarily attribute this discrepancy to FACTOR's limited four-option answer format. To demonstrate this, in \autoref{fig:ace_vs_acc_bear_factor} we plot \emph{ACE} against \emph{Accuracy} for FACTOR and for the 1:1 relations of BEAR with five sampled answer options, i.e. one more than on FACTOR. We find that matching the number of answer options across datasets recovers a nearly identical relationship between accuracy and  \emph{ACE}. Hence, for identical accuracies, we find similar \emph{ACE} values on BEAR and FACTOR. 
For example, \texttt{gpt2-large} achieves an accuracy of 48.97\% on BEAR with an \emph{ACE} of 0.33, while \texttt{opt-6.7b} has a comparable accuracy on FACTOR of 49.33\% and a nearly identical \emph{ACE} of 0.323. Similar parity is observed for \texttt{gpt2-xl} 
and \texttt{gpt2}, 
confirming that calibration is impacted by the distractor count.

\begin{table}[htp!]
\centering
\small
\begin{tabularx}{\linewidth}{l*{3}{>{\centering\arraybackslash}X}}
\toprule

\textbf{Model} &
\textbf{Acc.} &
$C_{\text{Margin}}$ &
$C_{\text{Average}}$ \\
\midrule

\texttt{opt-125m} (125M) & 15.6 & \textbf{0.186} & \textbf{0.130} \\

\texttt{roberta-base} (125M) & 16.0 & 0.291 & 0.162 \\

\midrule

\texttt{gpt2-medium} (355M) & 18.4 & \textbf{0.179} & \textbf{0.110} \\

\texttt{bert-base-cased} (109M) & 18.1 & 0.276 & 0.205 \\

\midrule

\texttt{gpt2-large} (812M) & 21.5 & \textbf{0.163} & \textbf{0.116} \\

\texttt{roberta-large} (355M) & 21.1 & 0.278 & 0.158 \\

\midrule

\texttt{opt-350m} (350M) & 19.0 & \textbf{0.177} & \textbf{0.127} \\

\texttt{bert-large-cased} (335M)  & 19.3 & 0.285 & 0.196 \\

\midrule



\texttt{gemma-3-270m-it (270M)}  & 15.5 & \textbf{0.224} & \textbf{0.116} \\
\texttt{ModernBERT-base (149M)}  & 16.8 & 0.289 & 0.152 \\

\midrule

\texttt{Qwen3.5-0.8B (800M)}  & 21.6 & \textbf{0.182} & \textbf{0.115} \\
\texttt{ModernBERT-large (335M)}  & 22.0 & 0.262 & 0.148 \\

\bottomrule

\end{tabularx}
\caption{\emph{Brier Scores} ($\downarrow$) for matched CLM/MLM pairs. Accuracy (in \%) is the average over all templates.}
\label{tab:brier_model_pairs}

\end{table}

\subsection{Comparison of CLMs and MLMs}
\label{sec:clms_vs_mlms}
To examine the differences in calibration between the two pre-training objectives, we matched CLMs and MLMs with similar accuracies. 
The goal was to minimize the influence of differing accuracies on calibration. The matched model pairs are listed in \autoref{tab:brier_model_pairs}, together with the \emph{Brier Scores} of the best-performing estimates $C_\text{Margin}$ and $C_\text{Average}$. 

\noindent \textbf{Better Calibration of CLMs.} Although MLMs with the same or even lower number of parameters can achieve a similar, and in some cases even higher accuracy, we find that CLMs are much better calibrated using most of our estimates (see \autoref{tab:brier_model_pairs}). Across all five model pairs, we observe substantial calibration differences, often reaching differences of approximately 0.1 to 0.2 for $C_{\text{Margin}}$ and $C_{\text{Average}}$. For $C_{\text{Margin}}$ and $C_{\text{Average}}$, all CLMs achieve a \emph{Brier Score} of 0.2 or lower, while none of the MLMs do. 

\noindent \textbf{Softmax MLM scores do not reflect model confidence.}
MLMs do not explicitly model a sentence-level log-likelihood during their pre-training phase and we rely on the pseudo log-likelihoods, as explained in \autoref{sec:bear-probe}. While this scoring is sufficient to distinguish between correct and incorrect answers for the BEAR probe, we believe that the absence of a well-defined sentence-level log-likelihood is the main reason behind the large calibration discrepancies between CLMs and MLMs.

\section{Experiment 2: Impact of Linguistic Confidence Expressions}

This experiment examines the impact of adding epistemic markers to the statements, by injecting both verbal and numerical confidence expressions. 

\noindent \textbf{Injections show inconsistent calibration effects.}
We find that verbal weakeners such as \emph{possibly} improve calibration more consistently than numerical weakeners such as \emph{“I’m 0\% confident...”}, which exhibit lower and more variable effects across models. One possible explanation is that such expressions are less common in the training data and therefore less reliably interpreted as signals of uncertainty. We also note that instruction-tuning does not systematically shift log-likelihoods in line with the intended meaning of the injections: for \texttt{gemma-3-4b-it} and \texttt{gemma-3-12b-it}, \emph{certainly} reduces overconfidence and thereby lowers \emph{ACE}.
\begin{figure}[t!]
    \centering
    \includegraphics[width=1\linewidth]{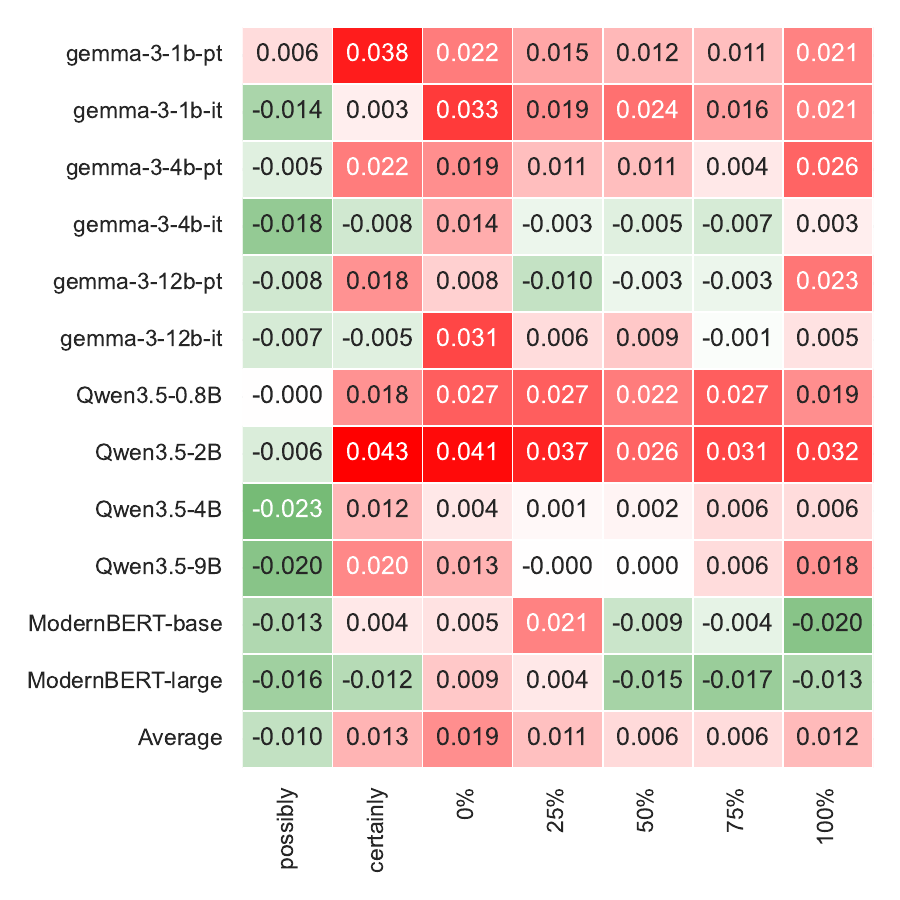}
    \caption{Difference in \emph{ACE} between the non-injected and injected templates. The x-axis shows the applied injection. Negative values (green) show an improvement.}
    \label{fig:aces_forced_epistemicity}
\end{figure}

\begin{figure}[t!]
    \centering
    \includegraphics[width=\linewidth]{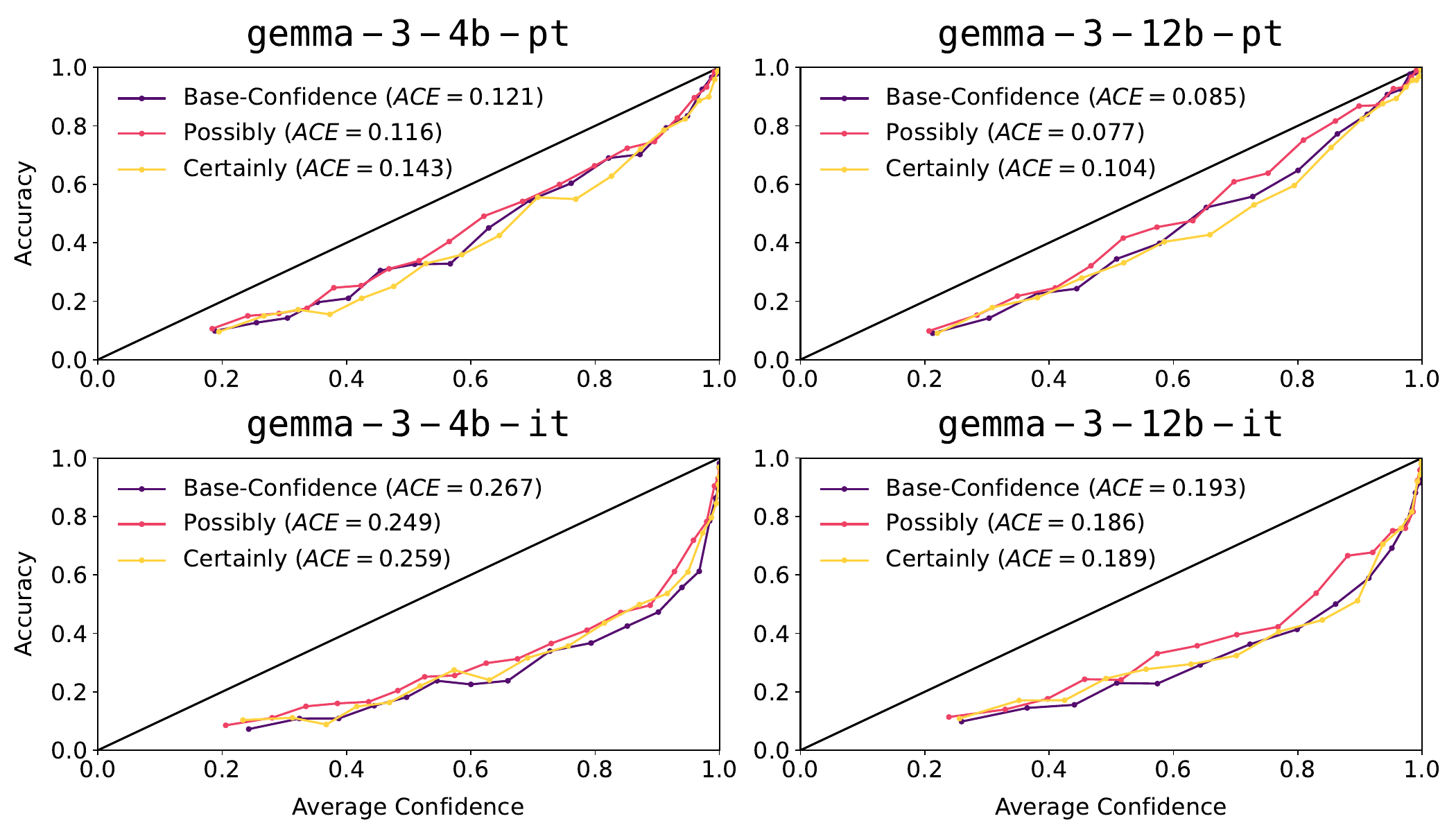}
    \caption{Calibration curves of $C_{\text{Base}}$ derived from the original and verbal injected templates.}
    \label{fig:ccurves_forced_epistemicity}
\end{figure}

\noindent \textbf{Verbal markers shift calibration curves.}
We confirm the expected positive impact of \emph{possibly} on \emph{ACE} by plotting calibration curves for the base and instruction-tuned variants of \texttt{gemma-3-4b} and \texttt{gemma-3-12b} (\autoref{fig:ccurves_forced_epistemicity}). Compared to the unaltered Base-Confidence, the injection of \emph{possibly} shifts the calibration curve closer to the identity line, thereby reducing \emph{ACE}. We also observe the somewhat unexpected finding for the instruction-tuned models, where the calibration curve under \emph{certainly} can lie above the curve of the non-injected baseline.

\section{Conclusion}

This study provides a comprehensive empirical analysis of the reliability of LMs' relational knowledge through the lens of calibration. Our evaluation of three distinct confidence estimation modalities—intrinsic confidence, paraphrase consistency and linguistic confidence—reveals that obtaining well-calibrated confidences for relational knowledge is challenging. Our findings strongly indicate that LMs lack a deeper understanding of their own knowledge boundaries; regardless of the pre-training objective and model, logit-based intrinsic confidences are consistently overconfident, a tendency more pronounced in MLMs. We demonstrate that the prediction margin can serve as a confidence signal superior to standard top-label confidence, and that calibration error significantly diminishes as the number of answer options increases.

Confidence estimates based on paraphrase consistency yield the best calibration, which reveals the LMs' high sensitivity to formulation changes. If this sensitivity is not incorporated in the confidence score, the LMs remain largely unreliable from a calibration perspective, especially for low-accuracy models. Since accounting for this sensitivity in practice requires costly prompt perturbations, the self-awareness of LMs in their factual knowledge remains limited.

\section*{Limitations}
Following are some limitations in our results that we want to address.

For our purposes, quantifying calibration with respect to the top answer is sufficient, since BEAR contains no N:M relations. However, a complete assessment of reliability would also account for such relations. Calibration must then be evaluated over all answer options, for example using the \emph{Brier Score} with one-hot-encoded ground-truth labels $\bm{y}_i$ or the Static Calibration Error \cite{nixon2020measuringcalibrationdeeplearning}.

Our calibration probing serves mainly as a diagnostic tool and is quite specific to closed-answer set scenarios. The calibration error values are therefore unlikely to transfer directly to more practical settings. Such scenarios typically lack predefined answer options, particularly across different formulations, necessary to derive paraphrase consistencies. Moreover, the predefined answer options limit our evaluation to output uncertainty, but unreliable responses can also stem from input uncertainty, such as vaguely formulated prompts.

\section*{Acknowledgments}
Elena Merdjanovska and Alan Akbik are supported by the Deutsche Forschungsgemeinschaft (DFG, German Research Foundation) under Germany’s Excellence Strategy – EXC 2002/2 “Science of Intelligence” – project number 390523135. Alan Akbik is supported by the Deutsche Forschungsgemeinschaft (DFG, German Research Foundation) under Emmy Noether grant “Eidetic Representations of Natural Language” (project number 448414230).


\bibliography{custom}

\appendix
\section{Overview of Models}
\label{sec:appendix-models}
\autoref{tab:models_list} lists all language models included in our experiments.

\begin{table}[htb!]
\centering
\small
\setlength{\tabcolsep}{4.2pt}
\begin{tabular}{lrrr}
\toprule
\textbf{Model} & \textbf{\# Param.} & \textbf{Objective} & \textbf{Accuracy} \\
\midrule

\texttt{gemma-2b}             & 2.0b  & CLM & 50.0 \\
\texttt{gemma-7b}             & 7.0b  & CLM & 62.4 \\

\midrule[0.3pt]

\texttt{gpt2}                 & 137M  & CLM & 12.8 \\
\texttt{gpt2-medium}          & 355M  & CLM & 18.4 \\
\texttt{gpt2-large}           & 812M  & CLM & 21.5 \\
\texttt{gpt2-xl}              & 1.6b  & CLM & 25.3 \\

\midrule[0.3pt]

\texttt{opt-125m}             & 125M  & CLM & 15.6 \\
\texttt{opt-350m}             & 350M  & CLM & 19.0 \\
\texttt{opt-1.3b}             & 1.3b  & CLM & 31.0 \\
\texttt{opt-6.7b}             & 6.7b  & CLM & 42.8 \\

\midrule

\texttt{gemma-3-270m (IT)} & 270M & CLM & 15.5 \\
\texttt{gemma-3-1b (IT)} & 1.0B & CLM & 23.6 \\
\texttt{gemma-3-4b (IT)} & 4.0B & CLM & 40.5 \\
\texttt{gemma-3-12b (IT)} & 12.0B & CLM & 53.9 \\

\midrule

\texttt{Llama-3.2-1B (IT)} & 1.0B & CLM & 37.6 \\
\texttt{Llama-3.2-3B (IT)} & 3.0B & CLM & 48.7 \\
\texttt{Llama-3.1-8B (IT)} & 8.0B & CLM & 62.2 \\

\midrule

\texttt{Qwen3.5-0.8B (IT)} & 800B & CLM & 21.6 \\
\texttt{Qwen3.5-2B (IT)} & 2.0B & CLM & 30.7 \\
\texttt{Qwen3.5-4B (IT)} & 4.0B & CLM & 42.6 \\
\texttt{Qwen3.5-9B (IT)} & 9.0B & CLM & 51.0 \\

\midrule[0.3pt]

\texttt{bert-base-cased}      & 109M  & MLM & 18.1 \\
\texttt{bert-large-cased}     & 335M  & MLM & 19.3 \\

\midrule[0.3pt]

\texttt{roberta-base}         & 125M  & MLM & 16.0 \\
\texttt{roberta-large}        & 355M  & MLM & 21.1 \\

\midrule[0.3pt]

\texttt{xlm-roberta-base}     & 279M  & MLM & 11.2 \\
\texttt{xlm-roberta-large}    & 561M  & MLM & 13.5 \\

\midrule

\texttt{ModernBERT-base} & 149M & MLM & 16.8 \\
\texttt{ModernBERT-large} & 335M & MLM & 22.0 \\ 
\bottomrule
\end{tabular}
\vspace{-0.5em}
\caption{
Models considered in this work. 
(IT) marks an instruction-tuned variant.
}
\label{tab:models_list}
\end{table}

\section{Extending the BEAR Templates}
\label{sec:appendix-extending-templates}
To better justify the paraphrase consistency estimates, we extended the number of templates from three to five per relation. To generate the additional templates, we utilized the free version of ChatGPT. We used the following prompt:
\bigskip
\begin{center}
\begin{minipage}{0.95\columnwidth}
\small
\texttt{As a research assistant, your task is to create an evaluation dataset to assess the relational knowledge of language models. You are provided with a specific relation label, and for each relation label, three semantically equivalent templates are already given, where '[X]' is a placeholder for the subject and '[Y]' for the object (answer). Your objective is to craft two additional semantically similar cloze sentence templates that capture the semantic meaning of the existing templates. Ensure that these new sentence templates are straightforward and devoid of superfluous elements.}

\smallskip

\texttt{For instance, given the following existing templates: ["[X] was authored by [Y].", "[X] is a written work by [Y].", "The author of [X] is [Y]."], an additional template might be: "[Y] wrote [X]."}

\smallskip

\texttt{Present your response in the JSON file format as you received it. Do not provide only the added templates, but include all five templates for each relation. You will be provided with the relation labels and the corresponding templates one by one in the following prompts.}
\end{minipage}
\end{center}

\section{Reduction Strategy Evaluation}
\label{sec:appendix-reduction-strategy}
 
We compare different strategies to reduce a set of token log-likelihoods to a single sentence-level log-likelihood.
Following are the four different strategies, i.e. variants of $C_{\text{Base}}$ (\autoref{def:base_confidence}):
\begin{enumerate}
	\item \emph{Sum}: summation of all tokens in the sequence: tokens corresponding to the object, subject, and  template.
    \item \emph{Mean}: averaging over all tokens in the sequence.
    \item \emph{Sum (A)}: summation over (answer) object tokens only.
	\item \emph{Mean (A)}: averaging over answer (object) tokens only.
\end{enumerate}
\begin{figure}[h!]
	\center
	\includegraphics[width=0.45\textwidth]{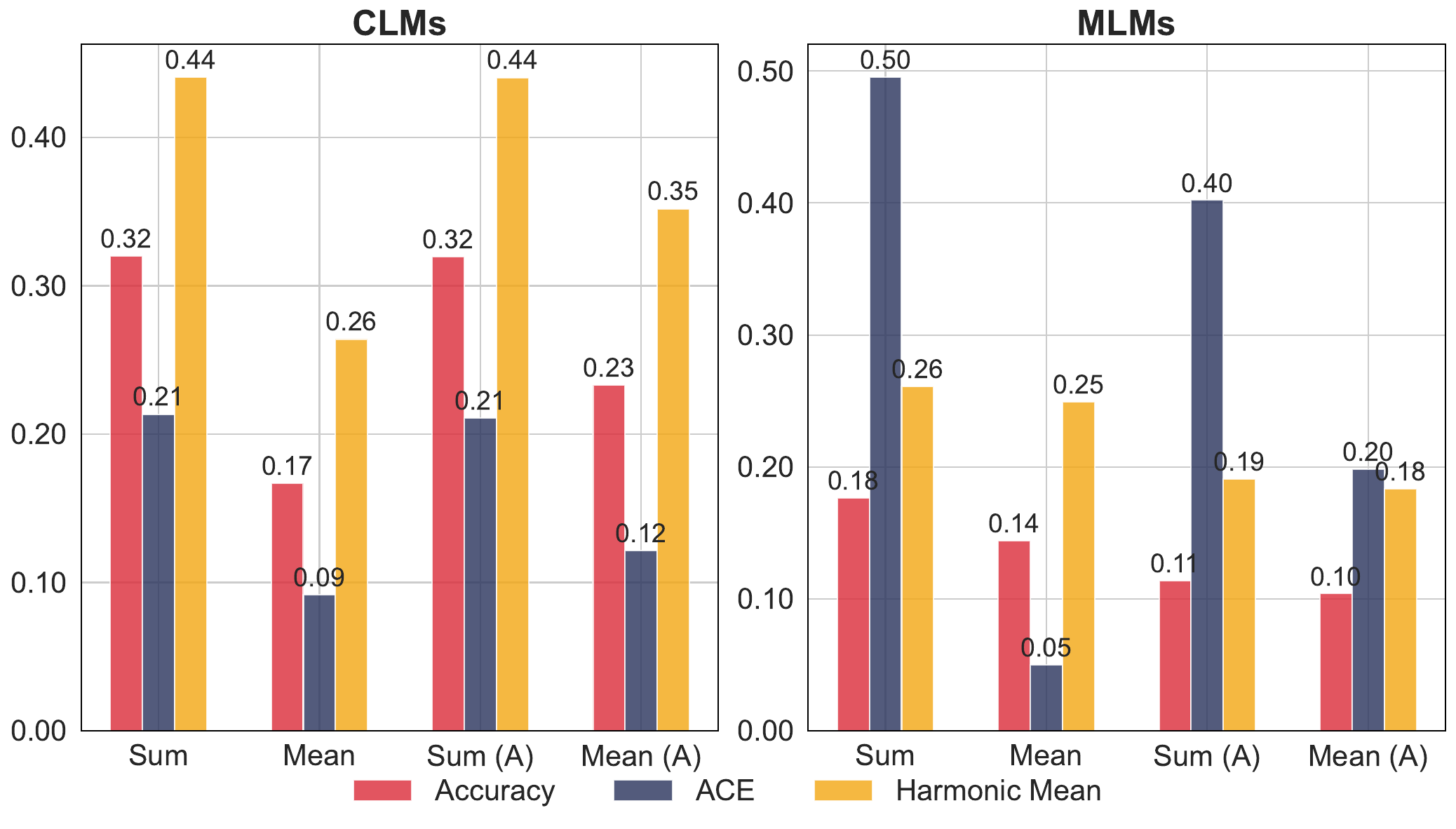}
	\caption[Barplot Reduction Strategies]{Performance of different token log-likelihood reduction strategies, measured by Accuracy, \emph{ACE} and harmonic mean. Higher harmonic mean is better. Scores are averaged across CLMs and MLMs.}
	\label{fig:reduction_strategies}
\end{figure}
We compare these strategies in terms of accuracy and \emph{ACE}. We additionally calculate their harmonic mean weighting accuracy and \emph{ACE} equally:
\begin{align*}
	H \coloneqq \frac{2 \cdot \text{ACC} \cdot (1 - \text{\emph{ACE}})}{\text{ACC} + (1 - \text{\emph{ACE}})} \in [0,1]
\end{align*}
where we subtracted the \emph{ACE} from 1 so that higher values reflect better overall performance.

\noindent In \autoref{fig:reduction_strategies}, we show the three metrics for each reduction strategy, averaged separately across CLMs and MLMs. We selected the full-sequence summation --- $Sum$ --- as our preferred approach, as it resulted in highest $H$ scores for both CLMs and MLMs. $Mean$ and \emph{Mean (A)} do result in better calibration scores than the summation variants, however this comes with a significant decrease in accuracy, making it overall less beneficial.

\begin{figure}[h!]
    \centering
    \includegraphics[width=1\linewidth]{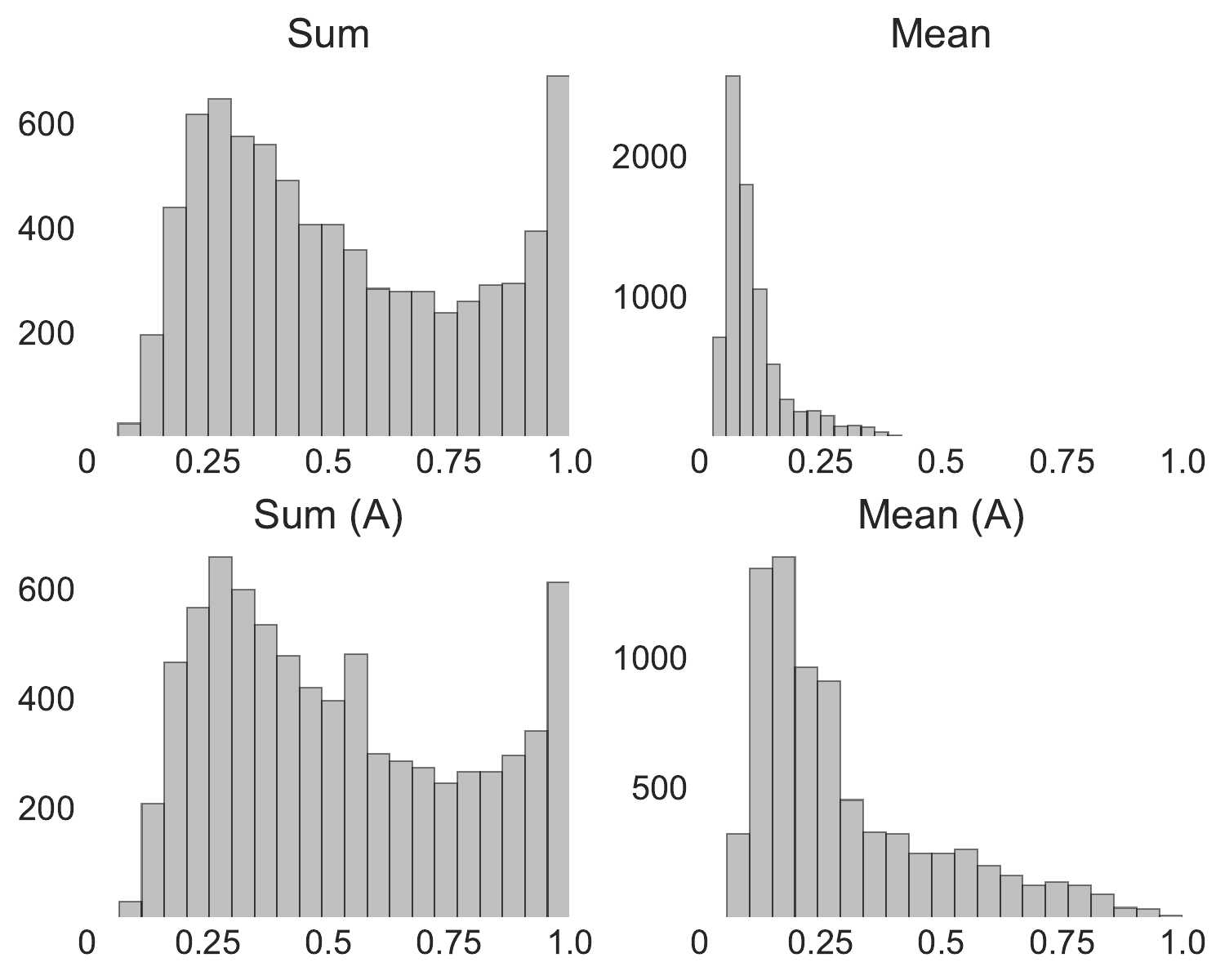}
    \caption{Distributions of $C_{\text{Base}}$ for \texttt{opt-1.3b} and the different reduction strategies.}
    \label{fig:conf_distributions_opt1_3b}
\end{figure}

\noindent Furthermore, as shown in \autoref{fig:conf_distributions_opt1_3b}, mean reduction yields strongly right-skewed confidence distributions, which can result in artificially low calibration errors, especially for low-accuracy models. For high-accuracy models, however, the effect is reversed: calibration error increases because confidences remain too low despite correct predictions. Moreover, the choice of reduction strategy itself limits the model’s ability to express high confidences. Under full-sequence mean reduction, the confidences of \texttt{opt-1.3b} do not exceed 0.5.

\section{Correlation between ACE and Accuracy}
\label{sec:appendix-correlation}
\begin{figure}[t]
	\center
	\includegraphics[width=\linewidth]{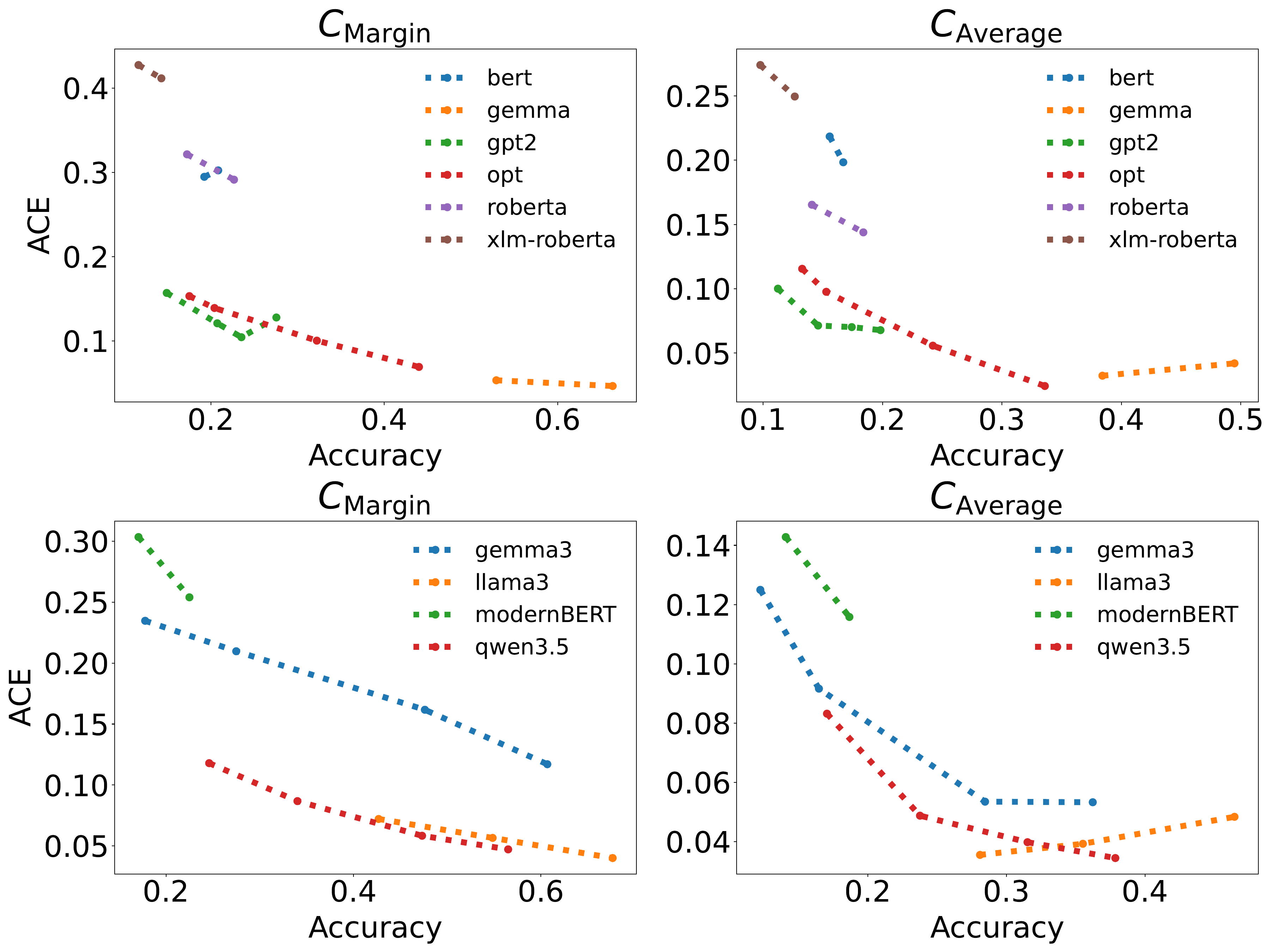}
	\caption[Correlation Plot Between Accuracy and \emph{ACE}]{\emph{ACE}-accuracy scatter plot, demonstrating negative correlation between the two metrics. The first row shows the \texttt{bert}, \texttt{gemma}, \texttt{gpt2}, \texttt{opt},  \texttt{roberta} and \texttt{xlm-roberta}  model families. The second row shows the \texttt{gemma-3}, \texttt{llama3}, \texttt{modern-bert} and  \texttt{qwen3.5} model families. }
	\label{fig:accuracy_vs_ace}
    \vspace{0.72cm}
\end{figure}

We find an overall negative correlation between accuracy and \emph{ACE}, exemplary shown in \autoref{fig:accuracy_vs_ace} for $C_{\text{Margin}}$ and $C_{\text{Average}}^{Min}$. Since larger models generally achieve higher accuracy, this result also implies a negative correlation between calibration error and model size. With some exceptions, this trend holds across pre-training objectives and model families.
From the perspective of overall trustworthiness, this is concerning: it suggests that gains in calibration primarily arise from higher accuracy rather than from well-aligned confidences.

\section{Selective Prediction Evaluation}
\begin{figure}[t]
	\center
    \includegraphics[width=\linewidth]{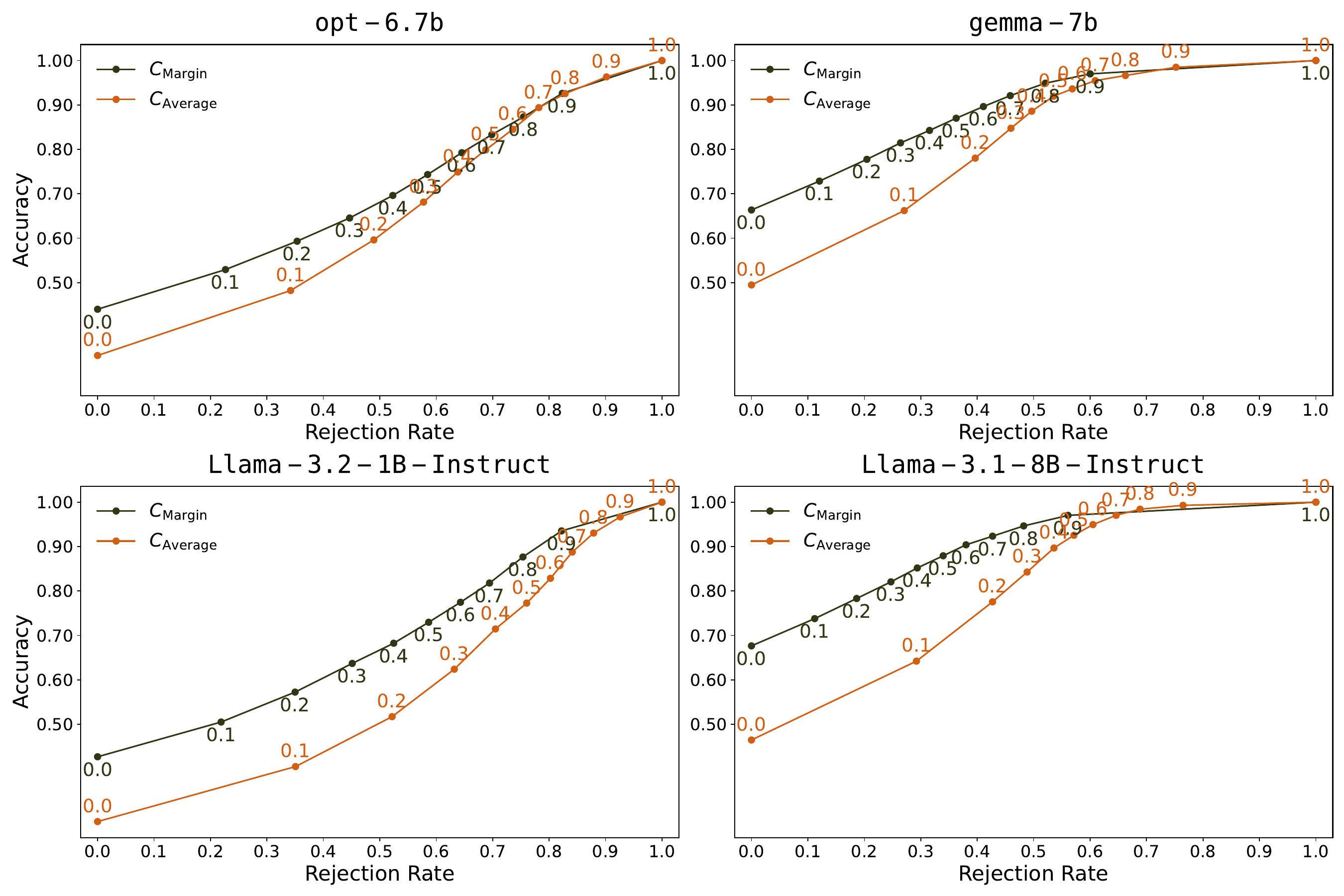}
	\caption[Accuracy Rejection Curves]{Accuracy Rejection Curves \protect\cite{nadeem2008accuracyrejectioncurves} (fraction of rejected answers versus accuracy among the non-rejected answers) for confidence thresholds of $0.1, \dots, 0.9$. For a given threshold, curves closer to the upper left indicate better performance.}
	\label{fig:arc_plot}
\end{figure}
We further evaluate the practical usability of the confidence estimates for selective prediction --- where we discard low-confidence data points in order to increase the accuracy on the remaining ones. \autoref{fig:arc_plot} shows the accuracy-rejection curves--- ploting the  share of discarded samples against the accuracy of the remaining ones --- of $C_{\text{Margin}}$ and $C_{\text{Average}}$ for selected models. These curves show that the confidence estimates are effective for selective prediction. For example, for \texttt{gemma-7b}, applying a confidence threshold of 0.1 to $C_{\text{Average}}$ increases accuracy by about 15 percentage points while rejecting less than 30\% of the dataset.

\section{Wiki-FACTOR Benchmark}
\label{sec:factor-benchmark}
While paraphrase consistency and linguistic confidence are specific to template-based probes such as BEAR, intrinsic confidence estimates apply to any multiple-choice setting. For a further benchmark, we implemented the estimates $C_{\text{Base}}$ and $C_{\text{Margin}}$ on the Wiki-FACTOR dataset \cite{muhlgay-etal-2024-generating}.
\newline
FACTOR defines a multiple-choice task in which each question consists of a multi-sentence context, a factual continuation and three non-factual distractors. In the original setup, the model’s answer is chosen as the completion with the highest mean log-probability. As shown on BEAR, mean reduction often results in very low confidence scores, which are not indicative of a well-calibrated model. We observe the same behavior on FACTOR: log-likelihoods do not span the full confidence interval and often differ only marginally, resulting in random guessing confidence splits. Combined with $C_{\text{Margin}}$, the vast majority of confidences would be close to zero. We therefore aggregate token log-likelihoods by summation as we did it for BEAR. 
\newline

\begin{table}[htb!]
\centering
\setlength{\tabcolsep}{3.5pt}
\small
\begin{tabularx}{\linewidth}{>{\raggedright\arraybackslash}X
>{\centering\arraybackslash}c
*{2}{>{\centering\arraybackslash}c}
*{2}{>{\centering\arraybackslash}c}}

\toprule
& & \multicolumn{2}{c}{\textbf{ACE ($\downarrow$)}} 
& \multicolumn{2}{c}{\textbf{Brier Score ($\downarrow$)}} \\
\cmidrule(lr){3-4} \cmidrule(lr){5-6}

\textbf{Model}
& $Accuracy$
& $C_{\text{Base}}$
& $C_{\text{Margin}}$
& $C_{\text{Base}}$
& $C_{\text{Margin}}$ \\
\midrule

\texttt{gemma-2b}
& 40.21
& \ACEcell{0.433} & \textbf{\ACEcell{0.387}}
& \BrierCell{0.457} & \textbf{\BrierCell{0.432}} \\

\texttt{gemma-7b}
& 16.50
& \ACEcell{0.740} & \textbf{\ACEcell{0.690}}
& \BrierCell{0.753} & \textbf{\BrierCell{0.720}} \\

\midrule[0.3pt]

\texttt{gpt2}
& 26.95
& \ACEcell{0.528} & \textbf{\ACEcell{0.482}}
& \BrierCell{0.545} & \textbf{\BrierCell{0.513}} \\

\texttt{gpt2-medium}
& 31.93
& \ACEcell{0.481} & \textbf{\ACEcell{0.438}}
& \BrierCell{0.504} & \textbf{\BrierCell{0.477}} \\

\texttt{gpt2-large}
& 35.74
& \ACEcell{0.448} & \textbf{\ACEcell{0.433}}
& \BrierCell{0.484} & \textbf{\BrierCell{0.468}} \\

\texttt{gpt2-xl}
& 37.81
& \ACEcell{0.427} & \textbf{\ACEcell{0.388}}
& \BrierCell{0.459} & \textbf{\BrierCell{0.439}} \\

\midrule[0.3pt]

\texttt{opt-125m}
& 29.53
& \ACEcell{0.505} & \textbf{\ACEcell{0.481}}
& \BrierCell{0.535} & \textbf{\BrierCell{0.515}} \\

\texttt{opt-350m}
& 33.03
& \ACEcell{0.475} & \textbf{\ACEcell{0.452}}
& \BrierCell{0.506} & \textbf{\BrierCell{0.485}} \\

\texttt{opt-1.3b}
& 41.88
& \ACEcell{0.395} & \textbf{\ACEcell{0.385}}
& \BrierCell{0.436} & \textbf{\BrierCell{0.427}} \\

\texttt{opt-6.7b}
& 49.33
& \ACEcell{0.323} & \textbf{\ACEcell{0.321}}
& \BrierCell{0.377} & \textbf{\BrierCell{0.376}} \\
\midrule
\texttt{gemma-3-270m} & 26.9 & \ACEcell{0.139} & \ACEcell{0.247} & \BrierCell{0.217} & \BrierCell{0.264}\\
\texttt{gemma-3-4b} & 44.9 & \ACEcell{0.164} & \ACEcell{0.402}& \BrierCell{0.275} & \BrierCell{0.416} \\
\texttt{gemma-3-12b} & 43.0 & \ACEcell{0.175} & \ACEcell{0.388} & \BrierCell{0.277} & \BrierCell{0.405} \\
\midrule
\texttt{Qwen3.5-2B} & 45.9 & \ACEcell{0.189} & \ACEcell{0.407} & \BrierCell{0.282} & \BrierCell{0.436}\\
\texttt{Qwen3.5-4B} & 53.9 & \ACEcell{0.243} & \ACEcell{0.490} & \BrierCell{0.309} & \BrierCell{0.504} \\
\texttt{Qwen3.5-9B} & 59.0 & \ACEcell{0.284} & \ACEcell{0.544} & \BrierCell{0.325} & \BrierCell{0.547} \\
\bottomrule
\end{tabularx}

\vspace{-0.3em}
\caption{
\emph{ACE} and \emph{Brier Score} for $C_{\text{Base}}$ and $C_{\text{Margin}}$, as well as accuracy on the Wiki-FACTOR \protect\cite{muhlgay-etal-2024-generating} dataset.
\textbf{Bold values} indicate the best-performing estimate per model.
\textbf{ACE:}
\protect\textcolor[rgb]{0.35,0.60,0.35}{\scalebox{1.5}{\textbullet}}~[0,0.05),
\protect\textcolor[rgb]{0.30,0.50,0.65}{\scalebox{1.5}{\textbullet}}~[0.05,0.10),
\protect\textcolor[rgb]{0.65,0.50,0.25}{\scalebox{1.5}{\textbullet}}~[0.10,0.20),
\protect\textcolor[rgb]{0.65,0.30,0.30}{\scalebox{1.5}{\textbullet}}~$\geq 0.20$;
\textbf{Brier Score:}
\protect\textcolor[rgb]{0.35,0.60,0.35}{\scalebox{1.5}{\textbullet}}~[0–0.10),
\protect\textcolor[rgb]{0.30,0.50,0.65}{\scalebox{1.5}{\textbullet}}~[0.10–0.20),
\protect\textcolor[rgb]{0.65,0.50,0.25}{\scalebox{1.5}{\textbullet}}~[0.20–0.25),
\protect\textcolor[rgb]{0.65,0.30,0.30}{\scalebox{1.5}{\textbullet}}~$\geq 0.25$. All \texttt{gemma-3} and \texttt{Qwen3.5} models are the instruction-tuned variants.
}
\label{tab:ace_brier_factor}
\end{table}

\noindent
\textbf{Higher calibration errors on FACTOR.} In \autoref{tab:ace_brier_factor} we present the \emph{ACEs} on FACTOR for $C_{\text{Base}}$ and $C_{\text{Margin}}$ for the CLMs. The first thing that stands out is that the calibration errors are significantly higher than the ones we obtained on BEAR. On BEAR we found an average \emph{ACE} of 0.213 for $C_{\text{Base}}$ and 0.107 for $C_{\text{Margin}}$ for the CLMs. In contrast, on FACTOR the \emph{ACEs} are 0.475 and 0.446, respectively. 
\newline
\textbf{$C_{\text{Margin}}$ also improves on FACTOR.} We further note that $C_{\text{Margin}}$ also consistently improves compared to $C_{\text{Base}}$ on FACTOR, but with much smaller magnitude. On BEAR we found an average improvement in \emph{ACE} among the CLMs of 0.106, while on FACTOR, the average improvement of $C_{\text{Margin}}$ is only 0.03. However, we find again a more similar performance if we restrict BEAR to only five answer options per instance, as we did it before. Then average improvement of $C_{\text{Margin}}$ is 0.07.

\section{Impact of Forced Epistemicity on Accuracy}
\label{sec:appendix-epistemicity-accuracy}

\begin{figure}[t!]
    \centering
    \includegraphics[width=1\linewidth]{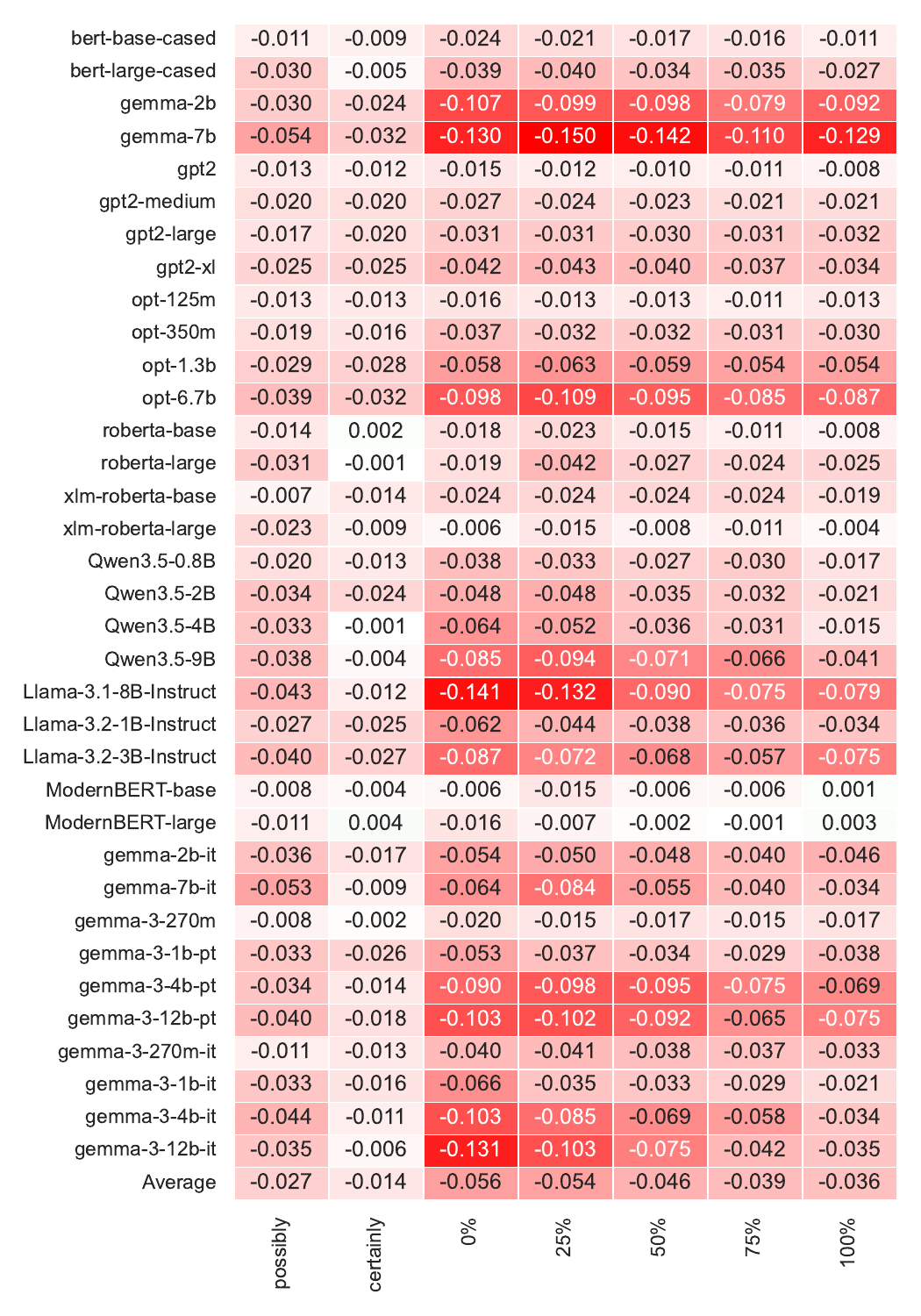}
    \caption{Difference in Accuracy between the non-injected and injected templates. Positive values show an improvement over the non-injected version.}
    \label{fig:acc_forced_epistemicity}
\end{figure}

While forced epistemicity can improve reliability for some models, we find that the injections significantly reduce accuracy, as shown in \autoref{fig:acc_forced_epistemicity}. For instance, \texttt{gemma-7b} shows a drop of 15 percentage points in accuracy for the injection of "I'm 25\% confident....", the largest decrease observed overall. The only LM where we observe a slight improvement compared to the unaltered template is \texttt{roberta-base}, where accuracy increased by 0.2 percentage points for the injection of "certainly". We also observe that larger models are more sensitive to these injections, particularly to numerical confidence expressions. 

\end{document}